\documentclass[lettersize,journal]{IEEEtran}
\usepackage{easyReview}
\usepackage{amsmath,amsfonts}
\usepackage{algorithmic}
\usepackage{algorithm}
\usepackage{array}
\usepackage[caption=false,font=normalsize,labelfont=sf,textfont=sf]{subfig}
\usepackage{textcomp}
\usepackage{stfloats}
\usepackage{url}
\usepackage{verbatim}
\usepackage{graphicx}
\usepackage{cite}
\usepackage{booktabs}
\usepackage{color}
\definecolor{gray}{gray}{0.5} 
\begin{document}

\title{HA-FGOVD: Highlighting Fine-grained Attributes via Explicit Linear Composition for Open-Vocabulary Object Detection}

\author{
  Yuqi Ma,
  Mengyin Liu,
  Chao Zhu,
  Xu-Cheng Yin,~\IEEEmembership{Senior Member,~IEEE,}

\thanks{This work was supported by National Natural Science Foundation of China under Grants 62072032 and 62076024, and National Science Fund for Distinguished Young Scholars under Grant 62125601.}
\thanks{Yuqi Ma, Mengyin Liu, Chao Zhu, and Xu-Cheng Yin are with School of Computer and Communication Engineering, University of Science and Technology Beijing, Beijing 100083, China (e-mails: yuqima@xs.ustb.edu.cn, blean@live.cn, chaozhu@ustb.edu.cn, xuchengyin@ustb.edu.cn).}
\thanks{Corresponding author: Chao Zhu.}}

\markboth{Journal of \LaTeX\ Class Files,~Vol.~14, No.~8, August~2021}%
{Shell \MakeLowercase{\textit{et al.}}: A Sample Article Using IEEEtran.cls for IEEE Journals}

\IEEEpubid{0000--0000/00\$00.00~\copyright~2021 IEEE}

\maketitle

\begin{abstract}
Open-vocabulary object detection (OVD) models are considered to be Large Multi-modal Models (LMM), due to their extensive training data and a large number of parameters. 
Mainstream OVD models prioritize object coarse-grained category rather than focus on their fine-grained attributes, e.g., colors or materials, thus failed to identify objects specified with certain attributes. 
However, OVD models are pretrained on large-scale image-text pairs with rich attribute words, whose latent feature space can represent the global text feature as a linear composition of fine-grained attribute tokens without highlighting them. Therefore, we propose in this paper a universal and explicit approach for frozen mainstream OVD models that boosts their attribute-level detection capabilities by highlighting fine-grained attributes in explicit linear space. Firstly, a LLM is leveraged to highlight attribute words within the input text as a zero-shot prompted task. 
Secondly, by strategically adjusting the token masks, the text encoders of OVD models extract both global text and attribute-specific features, which are then explicitly composited as two vectors in linear space to form the new attribute-highlighted feature for detection tasks, where corresponding scalars are hand-crafted or learned to reweight both two vectors.
Notably, these scalars can be seamlessly transferred among different OVD models, which proves that such an explicit linear composition is universal. Empirical evaluation on the FG-OVD dataset demonstrates that our proposed method uniformly improves fine-grained attribute-level OVD of various mainstream models and achieves new state-of-the-art performance. 
\end{abstract}


\section{Introduction}

\begin{figure}[!t]
	\centering
	\includegraphics[width=\columnwidth]{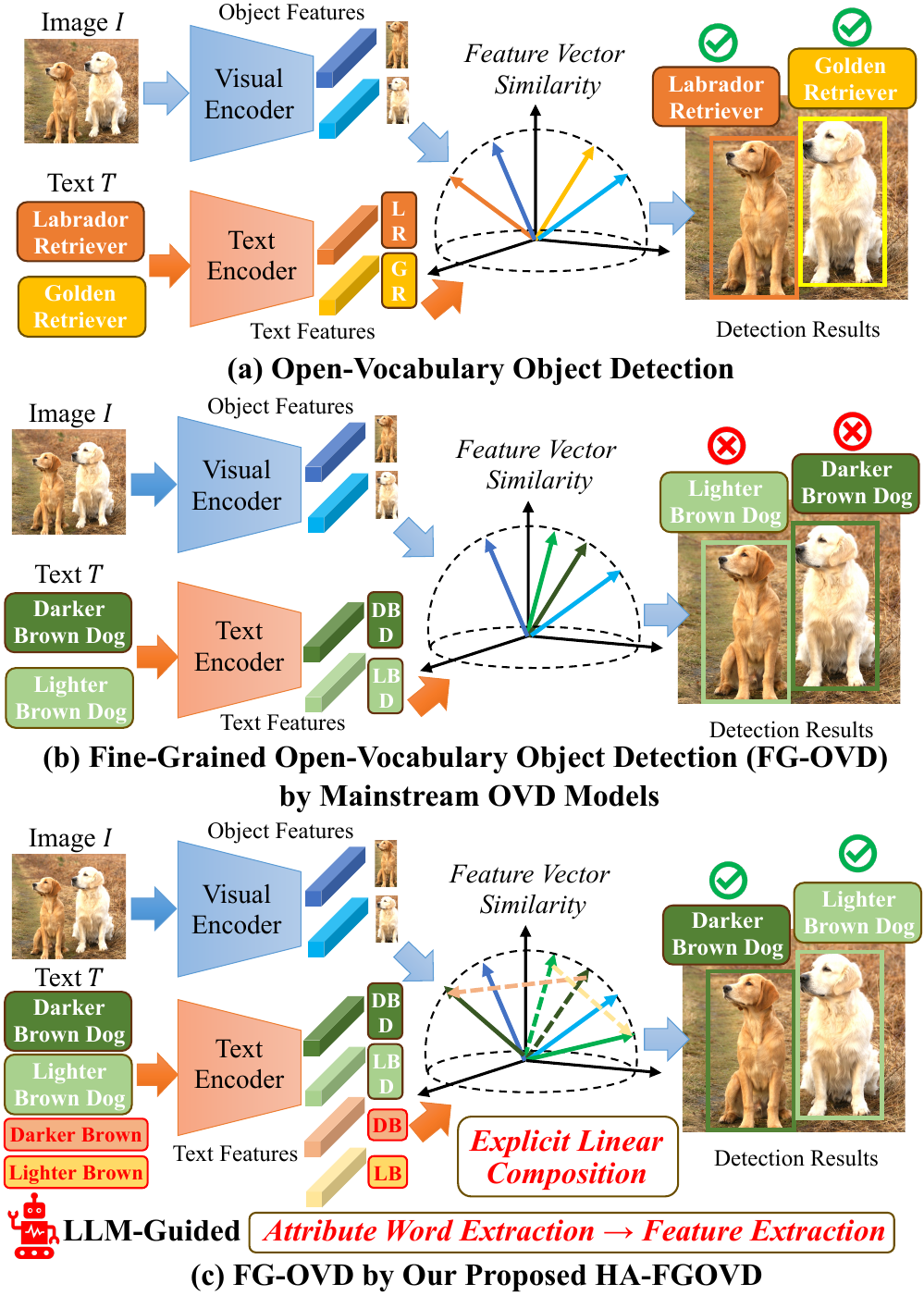}
	\caption{Difference between (a) open-vocabulary object detection for the fine-grained category names, (b) fine-grained open vocabulary object detection for the attribute-specific descriptions, and (c) our proposed HA-FGOVD method.}
	\label{fig_1}
\end{figure}

\IEEEPARstart{T}{he} Open-Vocabulary Object Detection (OVD) \cite{RN182} is based on \cite{RN263,RN157} or proposes an architecture of Large Multi-modal Models (LMM)\cite{RN141,RN210}, characterized by its extensive training data and a large number of parameters.
In contrast to classical object detectors limited to closed-set categories, OVD models detects objects further beyond the category scope of their training data \cite{RN138}. Although there has been a notable emergence of researches on OVD in recent years \cite{RN161, RN174, RN152}, major OVD models are both learned and evaluated on mainstream datasets with the typical labels as only a few category names. Therefore, they struggle in following natural user instruction with a more fine-grained attribute vocabulary than category names, which requires a stronger understanding of the fine-grained semantics of the language by aligning them with visual features. 

\IEEEpubidadjcol
To this end, recent studies have introduced a novel task, i.e., fine-grained open-vocabulary object detection (FG-OVD) \cite{RN149}. 
As illustrated in Fig. \ref{fig_1}, different from traditional closed-set fine-grained object detection (FG-OD) that focuses on inter-class variations (e.g., Labrador Retriever vs Golden Retriever). FG-OVD emphasizes extrinsic attributes (e.g., darker brown dog vs lighter brown dog) \cite{RN149}, which poses a challenge for OVD models. 
Moreover, the composition of various fine-grained attributes, e.g., color, patterns and materials, further enlarges such a fine-grained vocabulary that is not only broader than category names, but also closer to real-world application in natural language. 

Meanwhile, various works on FG-OVD have researched that insufficient emphasis on attribute words during the detection training phase leads to the inability of OVD models to recognize objects with characterized attributes \cite{RN149,RN254}. 
However, OVD models, either based on or proposed as large pretrained Vision-Language Models, leverage a vast array of image-text pairs \cite{RN141, RN148} enriched with attribute words. These models' latent feature spaces can represent global text features as a linear composition of fine-grained attribute tokens \cite{RN293}, while these attributes not being specifically highlighted within the OVD model.

To address the above issues, we have proposed a universal approach that explicitly enhances the FG-OVD capabilities of OVD models by highlighting fine-grained attributes in the explicit linear space. In a plug-and-play manner for frozen mainstream OVD models, powerful Large Language Model (LLM) \cite{RN202} is adopted for attribute word extraction, then text tokens are masked to guide the extraction of attribute-specific features by text encoder, which are then linearly composited with the global features for a new attribute-highlighted feature for better FG-OVD performance. The main contribution of our paper is summarized as:
\begin{enumerate}
\item{Firstly, Attribute Word Extraction is proposed to identify attribute words within the input text as a zero-shot prompted task based on a powerful LLM. }

\item{ Secondly, Attribute Feature Extraction strategically adjusting the token attention masks of text tokens, the text encoders of OVD models extract both global text and attribute-specific features. }

\item{ Thirdly, we propose Attribute Feature Enhancement based on the compositional property of text embedding. Global and attribute-specific features are fused as two vectors by explicit linear composition to form the new attribute-highlighted feature for further detection. }

\item{ Finally, experiments show the linear composition weight scalars can be learned or explicitly and seamlessly transferred among different OVD models. Evaluation on the FG-OVD dataset demonstrates that our proposed explicit and powerful approach significantly improves various mainstream OVD models and achieves new state-of-the-art performance.}
\end{enumerate}

\section{Related Works}
\label{relatedwork}
\subsection{Open Vocabulary Object Detection}

The task of OVD is designed to leverage rich vision-language semantics to detect objects of unseen categories during training. Hence, large vision-language models are pre-trained on large-scale image-text datasets with a rich vocabulary base. For instance, CLIP \cite{RN148}, which was trained on 400 million (image, text) pairs, has demonstrated significant potential in learning representations that are transferable to OVD models \cite{RN281} in various ways, including knowledge distillation \cite{RN145, RN162}, serving as an encoder for visual or textual data \cite{RN166, RN157}, and the generation of informative pseudo-labels \cite{RN175, RN155, RN152}. Due to limited regional text alignment capability of CLIP, some OVD models are directly trained on large-scale datasets for various tasks, such as object detection \cite{RN286, RN287, RN131}, image captioning \cite{RN285, RN284}, and phrase grounding \cite{RN287, RN288} etc., to enrich semantic understanding and improve the detection ability of OVD models. However, despite the inclusion of a rich attribute vocabulary in training datasets, recent studies \cite{RN149} have shown that mainstream OVD models perform poorly in detecting objects with fine-grained attributes. Under the mainstream evaluation protocol for only category names, OVD models suppress the fine-grained attributes that are explicitly composited in global text features\cite{RN255}. 

Consequently, the development of specialized datasets and methods for fine-grained open vocabulary detection is crucial for advancing the precision and granularity of OVD models.

\subsection{Fine-Grained Open Vocabulary Object Detection}

Traditional fine-grained object detection focuses on inter-class differences \cite{RN45, RN51, RN43, RN48}, such as distinguishing between Samoyed and Labrador dogs, and lack the capability to detect objects outside their training categories. In contrast, FG-OVD is more concerned with intra-class extrinsic attributes, such as ``brown dog'' or ``black dog'', offering greater generality. However, research on FG-OVD models is currently scarce. Most works focus on fine-grained categorization, e.g., DetCLIPv3 \cite{RN251} generates and trains on hierarchical category labels to enhance category-level fine-grained capabilities, while real-world objects are more complex with attribute-level granularity. 

Differently, this paper is dedicated to exploring attribute-level fine-grained open vocabulary detection methods. The study reveals that existing OVD models all incorporate text encoders based on the Transformer architecture \cite{RN112}, mainly based on CLIP text encoder architecture \cite{RN145, RN263, RN166, RN174, RN165, RN163, RN152, RN157}, and BERT architecture \cite{RN161, RN141, RN210}. Our proposed method can enhance text attribute features through linear combination to activate attribute features that exist but are suppressed by mainstream OVD models.

\subsection{Large Language Models as Auxiliary Tools} 

LLMs\cite{RN199, RN202}, due to their emergent comprehension and generation abilities to generate answers following user instructions, are increasingly being utilized to assist in various natural language processing tasks. For instance, OVD model DetCLIPv3 \cite{RN251} employs an LLM to annotate images with rich hierarchical object labels. Similarly, the FG-OVD dataset \cite{RN149} comprises both positive and negative captions generated by an LLM. In this paper, the capabilities of LLM are leveraged to extract attributes in input text, thereby aiding OVD models in focusing on and enhancing attribute features. This approach is designed to improve the sensitivity of OVD models to fine-grained attributes and their overall performance on fine-grained open-vocabulary object detection.



\begin{figure*}[!t]
  \centering
  \includegraphics[width=\textwidth]{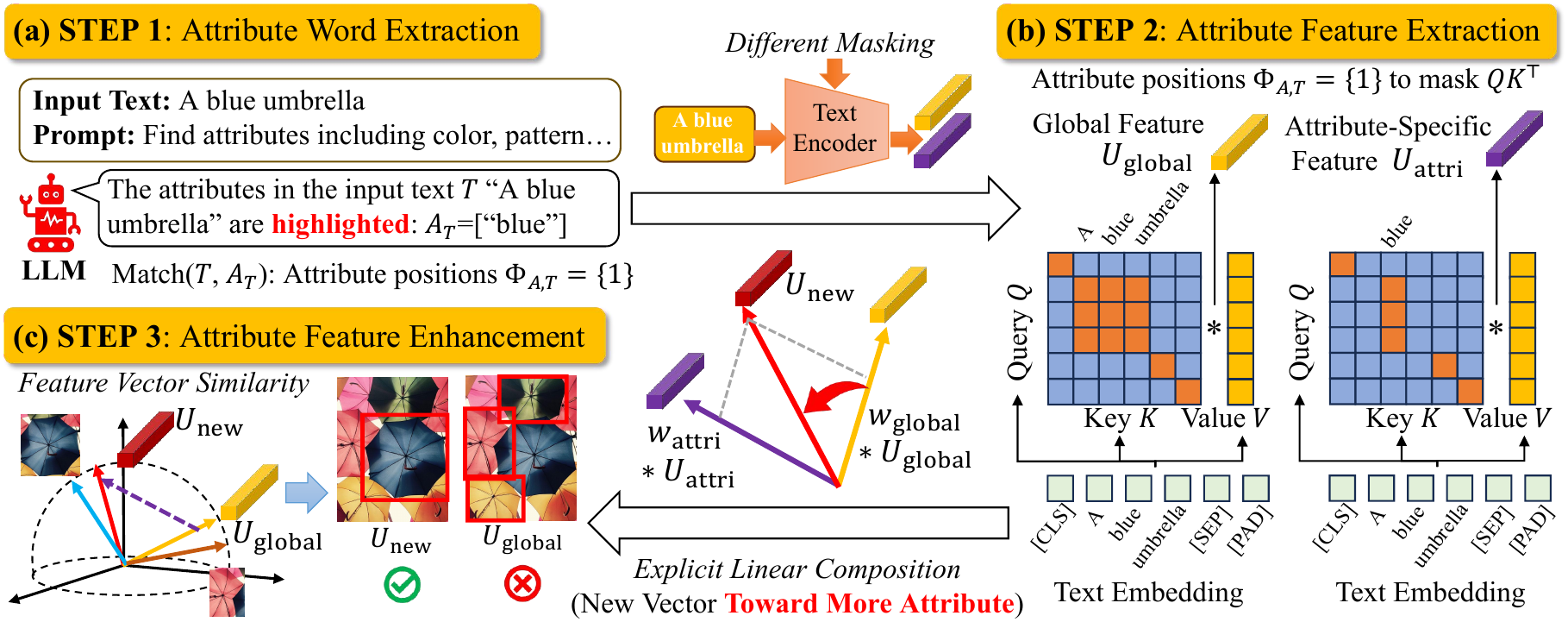}%
  \caption{The overall architecture of our proposed HA-FGOVD approach. (a) Firstly, a LLM follows the prompt to highlight the attributes in input text, which are then converted into attribute positions \(\Phi_{A,T}\). (b) Secondly, \(\Phi_{A,T}\) are  employed to mask the attention map \(QK^\top\) to obtain attribute specific feature \(U_\mathrm{attri}\). (c) Finally, explicit linear composition yields new feature \(U_\mathrm{new}\) from \(U_\mathrm{attri}\) and \(U_\mathrm{global}\) toward more attribute, which enhances the final detection results. }
  \label{fig_3}
  \end{figure*}

\section{Proposed Method}

As illustrated in Fig. \ref{fig_3}, we propose a universal approach to enhance the attribute-level OVD capabilities of mainstream models by highlighting fine-grained attributes in an explicit linear space. To amplify the fine-grained attribute features that exist but are suppressed in the frozen OVD models, our architecture consists of three key process: Attribute Words Extraction, Attribute Feature Extraction and Attribute Feature Enhancement. 

\subsection{Attribute Words Extraction}
\begin{figure}[!t]
	\centering
	\includegraphics[width=\columnwidth]{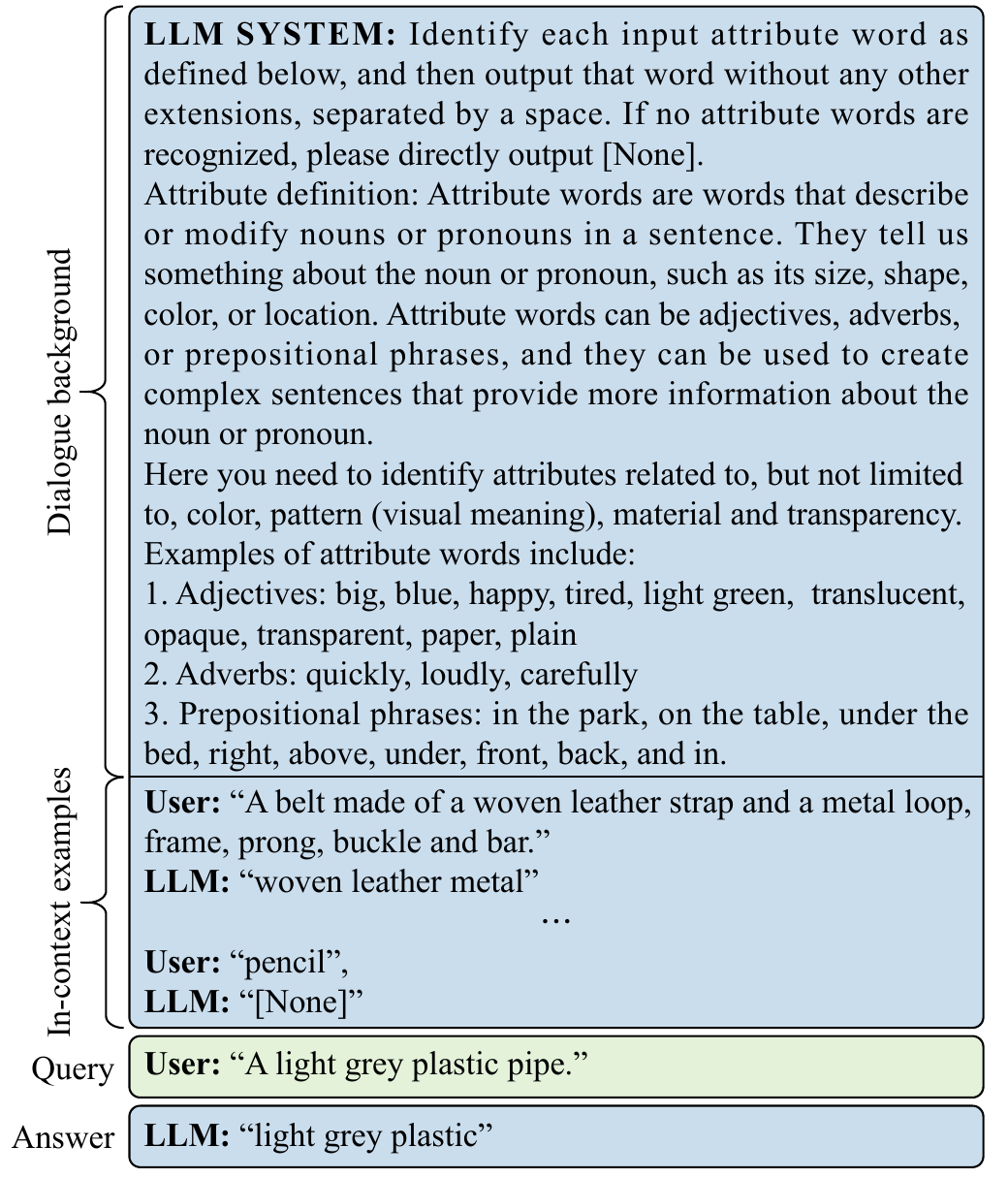}
	\caption{Attribute Words Extraction. The LLM is configured within the system message to establish a general dialogue background. This configuration defines the model's role as extracting attribute words and provides the definition of attribute words as well as the output format. In addition, 15 in-context examples are given to aid the LLM in comprehending the output rules, with the aim of improving the precision of attribute words extraction and reducing the risk of hallucinations.}
	\label{fig_llm}
\end{figure}

\begin{figure}[!t]
  \centering
  \includegraphics[width=\columnwidth]{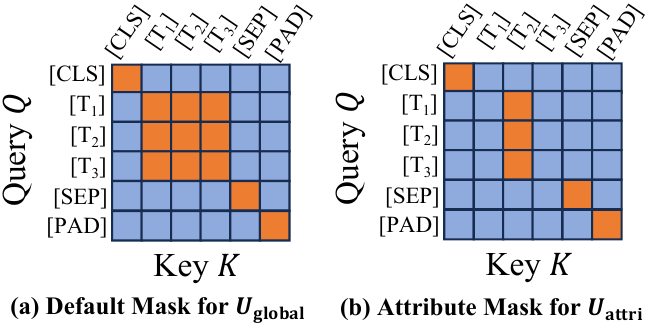}%
  \caption{2D attention masks in text encoder of BERT architecture. (a) Default mask for global feature \(U_\mathrm{global}\) and (b) attribute mask for attribute-specific feature \(U_\mathrm{attri}\). Take the 2\textsuperscript{nd} token ``[T\textsubscript{2}]'' as attribute for an example. }
  \label{fig_4}
\end{figure}

\begin{figure}[!t]
  \centering
  \includegraphics[width=\columnwidth]{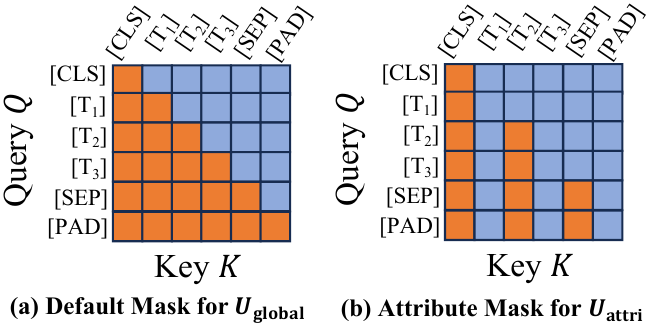}%
  \caption{2D attention masks in text encoder of CLIP architecture. (a) Default mask for global feature \(U_\mathrm{global}\) and (b) attribute mask for attribute-specific feature \(U_\mathrm{attri}\) . Take the 2\textsuperscript{nd} token ``[T\textsubscript{2}]'' as attribute for an example. }
  \label{fig_5}
\end{figure}
To assist the OVD models in focusing on attribute words, we employed the LLAMA2 \cite{RN202} LLM to extract attribute words from the input text. 
As shown in Fig. \ref{fig_llm}, to ensure the precision and standardization of the output, we configure the LLM within the system message to establish a general dialogue background which sets the model as the role of extracting the attribute words, and provides the definition of the attribute word as well as the output format. 
In addition, several in-context examples are supplied to aid the model in understanding the output rules. 
The specific output format is defined to include only attribute words, separated by spaces, without any other vocabulary, in the form of \([\textit{attribute1} \ \textit{attribute2} \ ... \ \textit{attributeN}  ]\). When no attribute word is detected in the statement, the output should be \([None]\). 

Given the prompt instruction \(P\) and input text \(T\), the LLM predicts the set of \(N\) attribute words \(A_T=\{a_i \in T\}\):
\begin{equation}
 A_T = \mathrm{LLM}(P,T) = \{a_i \in T~\vert~ 0 \leq i \leq \mathrm{N}\}
\end{equation}

Then the position of each attribute word is obtained by matching function with the input text \(T\):
\begin{equation}
  \begin{split}
  \Phi_{A,T} &= \mathrm{Match}(\mathrm{Tokenize}(T), \mathrm{Tokenize}(A_T))  \\
  &= \{ \phi_i \in [0,L] ~\vert~ 0 \leq i \leq \mathrm{N} \}
\end{split}
 \end{equation}

 \subsection{Attribute Feature Extraction}

 As mentioned in Section \ref{relatedwork}, all models we have surveyed include a text encoder structure and are based on variations of the Transformer architecture, specifically the CLIP \cite{RN148} or BERT \cite{RN270} models. 
 These models utilize token attention masks to guide the text encoder in extracting text information from specified positions when encoding input text. 
 Considering that the semantic information of attribute words depends on different contexts, for example, the contextual semantic information of ``blue'' in ``a blue dog'' and ``a blue umbrella'' is different, and in order to ensure consistency of the positional information between the global text features and attribute features, attribute features are not extracted from individual attribute words. 
 
 Instead, we have highlighted attribute semantic information by retaining the token attention masks of attribute and masking non-attribute tokens. 
 Subsequently, we combine the global text features with attribute features through a linear weighting composition to form an attribute-highlighted text feature, which is then utilized for object detection.
 
 The method is applied following the original models' token attention mask structures and making adaptive adjustments.

BERT architecture \cite{RN270} employs a vanilla self-attention mechanism which allows each element in the sequence to interact with all other elements within the sequence. Due to the model's special handling of attention masks for special tokens, such as the initial [CLS], separator [SEP], and padding [PAD] tokens (as illustrated in Fig. \ref{fig_4}), this algorithm operates only on the token attention masks between the [CLS] and [SEP] tokens.
 
Given an attention map \(Q K^{\top} \in \mathbb{R}^{(L+3) \times (L+3)}\), its default 2D attention mask \(M \in \{-\infty, 0\}^{L \times L}\) is derived as:
 \begin{equation}
 \bar{M} = \Psi \Psi^{\top} + \mathrm{diag}(1 - \Psi), \quad
M = [(1 - \bar{M}) \odot -\infty],
\end{equation}
which is further based on the 1D token mask as a vector \(\Psi=[\Psi_0, \Psi_1, \cdots, \Psi_{L+1}, \Psi_{L+2}]^\top\) with elements \(\Psi_i=1\) as text tokens and \(\Psi_i=0\) as special tokens [CLS], [PAD] and [SEP]:
\begin{equation}
  {\Psi_i} = \begin{cases}
  0,&{ i=0 \ ,\ i \ \textgreater \ L  }\\
  {1,}&{1 \le i \leq L} 
  \end{cases}.
\end{equation}

Therefore, \(\Psi\Psi^\top\) means that 2D attention is computed across the tokens of input text, while \(\mathrm{diag}(1 - \Psi)\) handles the special tokens at \(\Psi_i=0\) positions in \(1-\Psi\) as a Diagonal Matrix. 

With the attention mask \(M\), global feature \(U_\mathrm{global}\) is obtained via the self-attention mechanism with query \(Q\), key \(K\) and value \(V\) features of the text and special tokens as:
\begin{equation}
  U_\mathrm{global}= \mathrm{softmax}\left(\frac{QK^{\top}}{\sqrt{d_K}} + M\right)V,
\end{equation}
where each attention value added with \(-\infty\) in attention mask \(M\) yields \(e^{-\infty} = 0\) inside \(\mathrm{softmax}(\cdot)\) function for the lowest attention, and other value added with 0 

To keep attribute-specific tokens, we employ the attribute position \(\Phi_{A,T}\) from LLM as a new 1D token mask with elements \(\Theta_i\):
\begin{equation}
  {\Theta_i} = \begin{cases}
  0,&{ i  \in \Phi_{A,T} }\\
  {1,}&{ i \not\in \Phi_{A,T}} 
  \end{cases}
  \end{equation}

Attribute-specific feature \(U_{attri}\) is obtained similar to \(U_{global}\) via new \(M^{\ast}\) to obtain \(M^{\ast}\):
\begin{equation}
\bar{M}^* = \Theta \Psi^{\top} + \text{diag}(1 - \Psi)
\label{eq:GDINO}
\end{equation}

The 3D attention mask processing pseudocode is presented as Algorithm \ref{alg:alg11}. The \(bs\) and \(num\_text\_tokens\) are used for batch size and the number of text tokens.
The value \(True\) indicates that the token is not masked, and \(False\) indicates that it is masked.
\begin{algorithm}[t!]
  \caption{HF-FGOVD Algorithm Pipeline}\label{alg:alg11}
  \begin{algorithmic}
  \STATE 
  \STATE {\textbf{\textsc{FUNC}}}\ {\textsc{Match}}$(Token_\mathbf{cap},Token_\mathbf{attri})$
  
  \STATE \hspace{0.3cm}$ Position_\mathbf{attri}[bs,num\_text\_tokens] \gets  False$
  \STATE \hspace{0.3cm}$ \textbf{ for }  bs,(cap, attri) \ \textbf{in enum}(Token_\mathbf{cap},Token_\mathbf{attri}):  $
  
  \STATE \hspace{0.6cm}$ index \gets  [i \textbf{ if } cap_i = attri_j]$
  \STATE \hspace{0.6cm}$ \textbf{ if } model \textbf{ is } "BERT":  $
  
  \STATE \hspace{0.9cm}$ index \gets  index + [i \textbf{ if } cap_i = [\mathrm{PAD}]]$
  
  \STATE \hspace{0.6cm}$ Position_\mathbf{attri}[bs][index] \gets  True$

  \STATE \hspace{0.3cm}\textbf{return}  $Position_\mathbf{attri}$
  \STATE 
  \STATE {\textbf{\textsc{FUNC}}}\ {\textsc{Text\_Encoder}}$(Caption, Attribute, \theta')$ 
  \STATE \hspace{0.3cm}$ w_{\mathbf{global}},w_{\mathbf{attri}},\mathrm{bias} \gets \theta'$
  \STATE \hspace{0.3cm}$ Token_\mathbf{cap} \gets  \mathrm{Tokenize}(Caption)$
  \STATE \hspace{0.3cm}$ Mask_\mathbf{cap} \gets  \mathrm{Generate\_Mask} (Token_\mathbf{cap})$
  \STATE \hspace{0.3cm}$ Feature_\mathbf{cap} \gets  \mathrm{Attention} (Token_\mathbf{cap},Mask_\mathbf{cap})$
  \STATE \hspace{0.3cm}$ \textbf{ if }  Attribute \ \textbf{is not} \ None:  $
  \STATE \hspace{0.6cm}$ Token_\mathbf{attri} \gets  \mathrm{Tokenize} (Attribute)$
  \STATE \hspace{0.6cm}$ Position_\mathbf{attri} \gets  \mathrm{Match}  (Token_\mathbf{cap},Token_\mathbf{attri})$

  \STATE \hspace{0.6cm}$ Mask_\mathbf{attri} \gets  \text{Eq. \ref{eq:GDINO} or Eq. \ref{eq:OWL}}$

  \STATE \hspace{0.6cm}$ Feature_\mathbf{attri} \gets  \mathrm{Attention} (Token_\mathbf{cap},Mask_\mathbf{attri})$ 

  \STATE \hspace{0.6cm}$ Feature_{\mathbf{cap}} \gets  w_{\mathbf{global}} \times Feature_{\mathbf{cap}} +    $
  \STATE \hspace{3.0cm}$ w_{\mathbf{attri}} \times Feature_{\mathbf{attri}} + \text{bias}$
  \STATE \hspace{0.3cm}\textbf{return}  $Feature_\mathbf{cap}$
  \STATE
  
  \STATE {\textbf{\textsc{FUNC}}}\ {\textsc{Train}}$(Images, Captions, Prompt, GTs)$
  \STATE \hspace{0.3cm}$ \theta' \gets  w_{\mathbf{global}},w_{\mathbf{attri}},\mathrm{bias} $
  \STATE \hspace{0.3cm}$ \textbf{for}\ Image, Caption, GT\ \textbf{in}\ Images, Captions,GTs:$
  \STATE \hspace{0.6cm}$Attribute \gets \mathrm{LLM}(Prompt, Caption)$
  \STATE \hspace{0.6cm}$Feature_\mathbf{cap} \gets \mathrm{Text\_Encoder}(Caption,Attribute,\theta')$
  \STATE \hspace{0.6cm}$Feature_\mathbf{vis} \gets \mathrm{Visual\_Encoder}(Image)$
  \STATE \hspace{0.6cm}$Pred \gets \mathrm{Detection}(Feature_\mathbf{cap},Feature_\mathbf{vis})$
	
  \STATE \hspace{0.5cm} $\mathcal{L}_\mathbf{det} \gets \mathcal{L}_\mathbf{cls}(Pred_\mathbf{cls}, GT_\mathbf{cls}) + \mathcal{L}_\mathbf{reg}(Pred_\mathbf{reg}, GT_\mathbf{reg})$
  
  \STATE \hspace{0.5cm} $\theta' \gets \theta' - \eta \nabla_{\theta'} \mathcal{L}_\mathbf{det}$
  
  \STATE \hspace{0.2cm} $\textbf{return}\ w_{\mathbf{global}},w_{\mathbf{attri}},\mathrm{bias}$
	
  \end{algorithmic}
  \label{alg11}
  \end{algorithm}

CLIP architecture \cite{RN148} employs a lower triangular matrix causal attention mechanism, which is a special kind of self-attention mechanism. This mechanism introduces directionality into the sequence model, ensuring that when calculating the representation of the current element, only the positions before the current element are considered, without including those after it. 
As shown in Fig. \ref{fig_5}, unlike self-attention, causal attention does not perform special processing on the attention mask corresponding to the special token. 
Since [CLS] tokens and [SEP] tokens contain global and location information \cite{RN291, RN292}, these two special tokens are not masked when extracting attribute features. 
The [PAD] token is masked which has no practical meaning. 

Default 2D attention mask \(M\in\left\{-\infty,0\right\}^{L\times L}\) is derived as a Lower Triangle Matrix:
\begin{equation}
  { M_{i,j}} = \begin{cases}
  0,&{\ i \ \geq \ j  }\\
  {-\infty, }&{ \ i \ \textless \ j }
  \end{cases}
\end{equation}

Now given the attribute position \({\Phi}_{A,T}\) from LLM, we have new 2D mask \(M^\ast\):
\begin{equation}
  { M_{i,j}^\ast} = \begin{cases}
  0,&{ \ \  i \geq j\ and \ j \in {\Phi}_{A,T}  }\\
  {-\infty,}&{\ \ otherwise} 
  \end{cases}
  \label{eq:OWL}
\end{equation}

Both global feature \(U_{global}\) and attribute-specific feature \(U_{attri}\) are obtained similar to self-attention equation in BERT. 
The algorithm can also be represented by Algorithm \ref{alg11}.

\subsection{Attribute Feature Enhancement}

In the field of multimodality, embeddings of composite concepts can often be well-approximated as linear compositional structures\cite{RN293}, such as \(U(\text{rainy\ morning}) = U(\text{rainy}) + U(\text{morning})\). Leveraging the linear additivity of embeddings, we perform a weighted linear fusion of global text features and attribute features as two vectors, which can be mathematically represented as follows:
\begin{gather}
  U_\mathrm{global} = U_\mathrm{attri} + U_\mathrm{cate},  \nonumber\\
  U_{\mathrm{new}} = w_\mathrm{global} U_\mathrm{global} + w_\mathrm{attri} U_\mathrm{attri} \nonumber\\
  \Rightarrow U_\mathrm{new} =w_\mathrm{global}\left(U_\mathrm{attri}+U_\mathrm{cate}\right)+w_\mathrm{attri}U_\mathrm{attri} \nonumber\\
  \Rightarrow U_\mathrm{new}=\left(w_\mathrm{global}+w_\mathrm{attri}\right)U_\mathrm{attri}+w_\mathrm{global}U_\mathrm{cate}
\end{gather}
where \(U_\mathrm{global}\) and \(U_\mathrm{attri}\) are the vectors for global text and the attribute words, respectively, while \(w_\mathrm{global}\) and \(w_\mathrm{attri}\) denote their corresponding weight scalars. 
According to the reasoning formula, the new features \(U_\mathrm{new}\) are enhanced more towards \(U_\mathrm{attri}\), with $w_\mathrm{global} + w_\mathrm{attri}$ higher than \(w_\mathrm{global}\) for category features \(U_\mathrm{cate}\). Moreover, an extra \(\mathrm{bias}\) can be further applied to \(U_\mathrm{new}\) for a better learning of explicit linear composition:
\begin{equation}
  \label{eq:bias}
	U_\mathrm{new} = w_\mathrm{global} U_\mathrm{global} + w_\mathrm{attri} U_\mathrm{attri} + \mathrm{bias}.
\end{equation}

The \(U_\mathrm{new}\) signifies the attribute-highlighted embedding. \(U_\mathrm{new}\) is the output of the text encoder for detection tasks. The scaler weight triplets are represented as \((w_\mathrm{global},w_\mathrm{attri},\mathrm{bias})\).

  \section{Experiments}
  We report the performance of applying our method to three models of different architectures, which are fine-tuned in the FG-OVD \cite{RN149} training set and tested on this benchmark. 
  In ``Experimental Settings'' Section, FG-OVD dataset and evaluation protocol is introduced. 
  The three models are introduced in "Baseline Methods" Section.
  Then, we present the fine-tuning schemes and compare the fine-tuned FG-OVD performance with the baselines and previous works. 

  Since only three parameter scalars for the explicit linear composition are fine-tuned, they have strong transferability. Hence, we report the test performance after transferring the best weight scalars triplets of the three models to each other.
  
 \subsection{Experimental Settings}
 {\bf{FG-OVD benchmark:}}
 The FG-OVD dataset \cite{RN149} is a comprehensive dataset with eight distinct scenarios, divided into difficulty-based and attribute-based benchmarks. 
 Each subset consists of both positive and negative captions, which are essential for a balanced assessment. 
 The difficulty-based benchmarks adjust the negative captions to make the lower or higher distinction between positive and negative captions, with the Trivial, Easy, Medium, and Hard subsets representing increasing levels of challenge. In details, 
 negative captions of Trivial are arbitrarily selected from other objects, while Easy, Medium, and Hard are crafted by successively replacing 3, 2, and 1 attributes, in that order. 
 On the other hand, the attribute-based benchmarks enable a more focused evaluation of OVD models to detect objects with a specific kind of attributes types, including Color, Material, Pattern and Transparency. 
 
 {\bf{Evaluation Protocol:}}
 The Evaluation Protocol for this paper strictly adheres to the settings of FG-OVD dataset \cite{RN149}, employing dynamic vocabularies for both training and inference. This means that for each detected object \(i\), the model identifies a category from \(\{c_i^\mathrm{pos},c_{i,1}^\mathrm{neg},...,c_{i,N}^\mathrm{neg}\}\), 
 Here, \( c_i^{pos} \)signifies the positive description of the object \(i\), while \( c_{i,1}^{neg} \) represents the first negative caption for object \( i \), with a total of \( N \). 
 As depicted in the formula \cite{RN252}, traditional evaluation metrics do not penalize the presence of an incorrect prediction with higher confidence occurring at the correct location. 
 
 For instance, assuming a OVD model is tasked with detecting a category with a positive caption ``red house'' and a negative caption ``green house'', 
 if the model fails to comprehend the semantic context of the text, it may detect two bounding boxes at the location of the ground truth ``red house'' and assign both these two labels accordingly. 
 
 In this way, the OVD model could trick the evaluation metrics, achieving an accuracy of 0.5 and a recall of 1.0 for that category. And if this scenario is consistent across all categories tested, the final mean average precision (mAP) would also be 0.5: 
\begin{equation}
  Precision = \frac{TP}{TP + FP} = \frac{1}{1 + 1} = 0.5, 
\end{equation}
\begin{equation}
Recall = \frac{TP}{\mathrm{num}(GT)} = \frac{TP}{TP + FN} = \frac{1}{1+0} = 1,
\end{equation}
where \(TP\) denotes the count of instances correctly identified for a particular category, whereas \(FP\) and \(FN\) refer to the instances that are mistakenly classified as part of that category or exist in ground truth \(GT\) but not detected. 
\(\mathrm{num}(GT)\) signifies the overall number of ground-truth instances present in the image. To accurately assess the model's fine-grained open-vocabulary detection capabilities, a class-agnostic NMS \cite{RN149} is applied in post-processing ensure that only one prediction per location is retained, irrespective of the class label.

When testing on the benchmark, we follow the configuration from \cite{RN149}, selecting 5 negative captions (\(N\)=5) in the difficulty-based sub-datasets and 2 (\(N\)=2) in the attribute-based sub-datasets for each category.

\begin{table*}[!t]
	\setlength{\abovecaptionskip}{0.05cm}
	\centering

	\caption{Differences among Detic, Grouding DINO, and OWL-ViT models, including their overall architecture and text encoders.
		Text Embedding Flow means how textual features from text encoder are utilized for OVD. 
		ViEmbed means visual embedding.\\
	}
	\label{tab:tablearch}
	\begin{tabular}{*{9}{c}}
		\toprule
		Model   & Overall Architecture & Text Encoder Type  & Text Encoder Attention Type & Text Embedding Flow  \\
			\midrule
			Detic          & Two-stage  & CLIP (pretrained)   & Causal-Attention  & Replace the classifier weights     \\
			Grounding DINO  & End-to-end & BERT (pretrained)    & Self-Attention  & Calculate similarity after cross-fused with the ViEmbed  \\
			OWL-ViT        & End-to-end & CLIP (fine-tuned)  & Causal-Attention   & Calculate similarity with ViEmbed\\
			
			\bottomrule
			
		\end{tabular}
	\end{table*}

\subsection{Baseline Methods}
As shown in Table \ref{tab:tablearch}, the method proposed in this paper is applied to three distinct architecture-based OVD models: Detic \cite{RN157}, Grounding DINO \cite{RN141} and OWL-ViT \cite{RN263}.

Detic \cite{RN157} is trained jointly with image classification and object detection datasets to respectively expand detector vocabulary and enhance the classification and regression ability of the model. A mainstream two-stage object detection framework \cite{RN273, RN275} is employed with the classification weights \(W\) replaced by CLIP embeddings of class names to transform a conventional detector into an open-vocabulary detector.

Grounding DINO\cite{RN141}, a Referring Expression Comprehension (REC) model, is capable of detecting any object as per the user's input of categories or indicative expressions. It enhances the closed-set Transformer-based detector, DINO, by incorporating multi-phase vision-language modality fusion. Furthermore, the text encoder component of Grounding DINO \cite{RN141} is based on a pre-trained BERT model.

OWL-ViT \cite{RN263} is an end-to-end vision Transformer (ViT) architecture that replaces the fixed classification layer weights with class-name embeddings derived from a text model. This text model is fine-tuned based on the CLIP \cite{RN148} text encoder.

OWL \cite{RN263} and Detic \cite{RN157} both utilize a text encoder based on the CLIP \cite{RN148} architecture, and their token attention mask structures are similar. Specific numerical values are used to represent the degree of encoding importance, typically using an extremely small value \(-\infty\) to denote no encoding and 0 for encoding, while they are marked by False and True in OWL model and then further converted into \(-\infty\) and 0, correspondingly.

\subsubsection*{\bf Model Architectural Settings} Implementation details of the baseline OVD models evaluated in this paper adhere to these settings following the paper of FG-OVD dataset \cite{RN149}: 
\begin{itemize}
\item{OWL-ViT: ViT L/14 backbone version.}
\item{Detic: \texttt{Detic\_LCOCOI21k\_CLIP\_SwinB\_896b32\_\\4x\_ft4x\_max-size}, i.e., Swin-B backbone version with ImageNet-21K pre-training.}
\item{Grounding Dino: Swin-T backbone version, i.e., the GroundingDINO-T configuration.}
\end{itemize}

\begin{table*}[!t]
  \setlength{\abovecaptionskip}{0.05cm}
  \centering

  \caption{ FG-OVD statistics for training and validation sets. \\
  With neg. caption indicates that the category contains negative captions.}
  \label{tab:table1}
  \begin{tabular}{*{9}{c}}
    \toprule
    Datasets & Hard & Medium & Easy & Trivial & Color & Material & Pattern & Transp. \\

    \midrule
    training set   & 27684 & 27684 & 27684 & 27684 & 27684 & 27684 & 27684 & 27684 \\
    with negative caption & 27683 & 24513 & 14178 & 27683 & 25215 & 25751 & 6020 & 2326 \\

    \midrule
    validation set & 1444 & 1270 & 761 & 1444 & 1326 & 1321 & 322 & 119 \\
    with negative caption & 1444 & 1270 & 761 & 1444 & 1326 & 1321 & 322 & 119 \\

    \bottomrule

  \end{tabular}
    \end{table*}

\begin{table*}[!t]
  \setlength{\abovecaptionskip}{0.05cm}
  \centering
  \caption{ MAP evaluation results on benchmark (\%). Average is averaged over the 8 sub-datasets. MAP from OWL-ViT(B/16) to CORA comes from previous work \cite{RN149}. Other mAP results are based on the experiments and reproduction (*) from our researches.}
  \label{tab:table2}
  \begin{tabular}{cllllllll|l}
    \toprule
    Detector & Hard & Medium & Easy & Trivial & Color & Material & Pattern & Transp. & Average\\

    \midrule
    OWL-ViT(B/16)   & 26.2 & 39.8 & 38.4 & 53.9 & 45.3 & 37.3 & 26.6 & \textbf{34.1} & 37.7 \\
    OWL-ViT(L/14)   & 26.5 & 39.3 & 44.0 & 65.1 & 43.8 & 44.9 & 36.0 & 29.2 & 41.1 \\
    OWLv2(B/16) & 25.3 & 38.5 & 40.0 & 52.9 & 45.1 & 33.5 & 19.2 & 28.5 & 35.4\\
    OWLv2(L/14) & 25.4 & 41.2 & 42.8 & 63.2 & \textbf{53.3} & 36.9 & 23.3 & 12.2 & 37.3\\
    Detic       & 11.5 & 18.6 & 18.6 & 69.7 & 21.5 & 38.8 & 30.1 & 24.6\footnotemark & 29.3  \\
    ViLD        & 22.1 & 36.1 & 39.9 & 56.6 & 43.2 & 34.9 & 24.5 & 30.1 & 35.9 \\ 
    Grounding DINO & 16.6 & 27.9 & 30.1 & 62.7 & 41.0 & 30.2 & 31.2 & 25.4 & 33.1\\
    CORA        & 13.8 & 20.0 & 20.4 & 35.1 & 25.0 & 19.3 & 22.0 & 27.9 & 22.9\\
    \midrule
    Detic*       & 11.5 & 18.6 & 18.8 & 69.7 & 21.6 & 38.8 & 30.3 & 24.8 & 29.3 \\
    + HA-FGOVD  & 12.2 \textcolor{red}{(+0.7)} & 22.0 \textcolor{red}{(+3.4)} & 19.8 \textcolor{red}{(+1.0)} & \textbf{70.3} \textcolor{red}{(+0.6)} & 23.7 \textcolor{red}{(+2.1)} & 41.0 \textcolor{red}{(+2.2)} & 32.1 \textcolor{red}{(+1.8)} & 25.5 \textcolor{red}{(+0.7)} & 30.8 \textcolor{red}{(+1.5)} \\
    \midrule
    Grounding DINO* & 17.0 & 28.4 & 31.0 & 62.5 & 41.4 & 30.3 & 31.0 & 26.2 & 33.5\\
    + HA-FGOVD  & 19.2 \textcolor{red}{(+2.2)} & 32.3 \textcolor{red}{(+3.9)} & 34.0 \textcolor{red}{(+3.0)} & 62.2 \textcolor{green}{(-0.3)} & 41.5 \textcolor{red}{(+0.1)} & 33.0 \textcolor{red}{(+2.7)} & 32.1 \textcolor{red}{(+1.1)} & 29.2 \textcolor{red}{(+3.0)} & 35.4 \textcolor{red}{(+1.9)}\\
    \midrule
    OWL-ViT(L/14)*   & 26.6 & 39.8 & 44.5 & 67.0 & 44.0 & 45.0 & 36.2 & 29.2 & 41.5\\
    + HA-FGOVD  & \textbf{31.4} \textcolor{red}{(+4.8)} & \textbf{46.0} \textcolor{red}{(+6.2)} & \textbf{50.7} \textcolor{red}{(+6.2)} & 67.2 \textcolor{red}{(+0.2)} & 48.4 \textcolor{red}{(+4.4)} & \textbf{48.5} \textcolor{red}{(+3.5)} & \textbf{38.0}  \textcolor{red}{(+1.8)} & 32.7 \textcolor{red}{(+3.5)} & \textbf{45.4} \textcolor{red}{(+3.9)}\\

    \bottomrule

  \end{tabular}
    \end{table*}

\begin{table*}[!t]
	\setlength{\abovecaptionskip}{0.05cm}
	\centering
	\caption{ Transfer result of weight scalar triplets (\%). Average is averaged over the 8 sub-datasets. The \textbf{Bolden} represents the highest mAP on the same model architecture.
		(\(\leftarrow Training\)) mean the weight scalar triplets is from training.  (\(\leftarrow Model\)) means to apply the best weight scalar triplets of the Model to the baseline. The best weight scalar triplets \((w_{global}, w_{attri}, bias)\) for each model is: Detic(0.434, 0.244, 0.250), Grounding DINO(0.565, 0.264, -0.178), OWL-ViT(0.772, 0.248, -0.080). The bias is set to 0 when transferring to another model.} 
	\label{tab:table3}
	\begin{tabular}{cllllllll|l}
		\toprule
		Detector & Hard & Medium & Easy & Trivial & Color & Material & Pattern & Transp. & Average\\
		
			\midrule
			Detic       & 11.5 & 18.6 & 18.8 & 69.7 & 21.6 & 38.8 & 30.3 & 24.8 & 29.3 \\
			\(\leftarrow\) Training  & \textbf{12.2} \textcolor{red}{(+0.7)} & \textbf{22.0} \textcolor{red}{(+3.4)} & \textbf{19.8} \textcolor{red}{(+1.0)} & \textbf{70.3} \textcolor{red}{(+0.6)} & \textbf{23.7} \textcolor{red}{(+2.1)} & \textbf{41.0} \textcolor{red}{(+2.2)} & \textbf{32.1} \textcolor{red}{(+1.8)} & 25.5 \textcolor{red}{(+0.7)} & \textbf{30.8} \textcolor{red}{(+1.5)} \\
			\(\leftarrow\) Grounding DINO  & 12.0 \textcolor{red}{(+0.5)} & 21.4 \textcolor{red}{(+2.8)} & 18.6 \textcolor{green}{(-0.2)} & 68.2 \textcolor{green}{(-1.5)} & 21.8 \textcolor{red}{(+0.2)} & 39.5 \textcolor{red}{(+0.7)} & 31.6 \textcolor{red}{(+1.3)} & 26.4 \textcolor{red}{(+1.6)} & 29.9 \textcolor{red}{(+0.6)} \\
			\(\leftarrow\) OWL-ViT(L/14)  & 11.9 \textcolor{red}{(+0.4)} & 21.0 \textcolor{red}{(+2.4)} & 18.8 (+0.0) & 69.8 \textcolor{red}{(+0.1)} & 22.5 \textcolor{red}{(+0.9)} & 40.5 \textcolor{red}{(+1.7)} & 31.9 \textcolor{red}{(+1.6)} & \textbf{26.6} \textcolor{red}{(+1.8)} & 30.4 \textcolor{red}{(+1.1)} \\
			
			\midrule
			Grounding DINO & 17.0 & 28.4 & 31.0 & 62.5 & 41.4 & 30.3 & 31.0 & 26.2 & 33.5\\
			\(\leftarrow\) Training  & \textbf{19.2} \textcolor{red}{(+2.2)} & \textbf{32.3} \textcolor{red}{(+3.9)} & \textbf{34.0} \textcolor{red}{(+3.0)} & 62.2 \textcolor{green}{(-0.3)} & \textbf{41.5} \textcolor{red}{(+0.1)} & \textbf{33.0} \textcolor{red}{(+2.7)} & \textbf{32.1} \textcolor{red}{(+1.1)} & 29.2 \textcolor{red}{(+3.0)} & \textbf{35.4} \textcolor{red}{(+1.9)}\\
			\(\leftarrow\) Detic  & 18.3 \textcolor{red}{(+1.3)} & 30.2 \textcolor{red}{(+1.8)} & 32.6 \textcolor{red}{(+1.6)} & \textbf{62.9} \textcolor{red}{(+0.4)} & 41.3 \textcolor{green}{(-0.1)} & 31.5 \textcolor{red}{(+1.2)} & 30.4 \textcolor{green}{(-0.6)} & 27.7 \textcolor{red}{(+1.5)} & 34.4 \textcolor{red}{(+0.9)} \\
			\(\leftarrow\) OWL-ViT(L/14)  & 19.0 \textcolor{red}{(+2.0)} & 31.1 \textcolor{red}{(+2.7)} & 33.5 \textcolor{red}{(+2.5)} & 62.7 \textcolor{red}{(+0.2)} & 41.3 \textcolor{green}{(-0.1)} & 32.1 \textcolor{red}{(+1.8)} & 31.7 \textcolor{red}{(+0.7)} & \textbf{30.0} \textcolor{red}{(+3.8)} & 35.2 \textcolor{red}{(+1.7)} \\
			
			\midrule
			OWL-ViT(L/14)   & 26.6 & 39.8 & 44.5 & 67.0 & 44.0 & 45.0 & 36.2 & 29.2 & 41.5\\
			\(\leftarrow\) Training  & 31.4 \textcolor{red}{(+4.8)} & 46.0 \textcolor{red}{(+6.2)} & 50.7 \textcolor{red}{(+6.2)} & \textbf{67.2} \textcolor{red}{(+0.2)} & 48.4 \textcolor{red}{(+4.4)} & 48.5 \textcolor{red}{(+3.5)} & 38.0  \textcolor{red}{(+1.8)} & \textbf{32.7} \textcolor{red}{(+3.5)} & \textbf{45.4} \textcolor{red}{(+3.9)}\\
      \(\leftarrow\) Detic  & 30.2 \textcolor{red}{(+3.6)} & 44.1 \textcolor{red}{(+4.3)} & 47.2 \textcolor{red}{(+2.7)} & \textbf{67.2} \textcolor{red}{(+0.2)} & 47.7 \textcolor{red}{(+3.7)} & 46.8 \textcolor{red}{(+1.8)} & \textbf{38.3} \textcolor{red}{(+2.1)} & 30.6 \textcolor{red}{(+1.4)} & 44.0 \textcolor{red}{(+2.5)} \\
			\(\leftarrow\) Grounding DINO  & \textbf{32.4} \textcolor{red}{(+5.8)} & \textbf{48.1} \textcolor{red}{(+8.3)} & \textbf{51.5} \textcolor{red}{(+7.0)} & 66.9 \textcolor{green}{(-0.1)} & \textbf{49.0} \textcolor{red}{(+5.0)} & \textbf{48.9} \textcolor{red}{(+3.9)} & 35.8 \textcolor{green}{(-0.4)} & 29.3 \textcolor{red}{(+0.1)} & 45.2 \textcolor{red}{(+3.7)} \\
			\bottomrule
			
		\end{tabular}
	\end{table*}

\subsection{Fine-Tuning Schemes}
As depicted in the Table \ref{tab:table1}, upon statistical analysis of the trainable data within each subsets of the FG-OVD training set, all 8 subsets contain 27,684 category annotations. However, a significant portion of the positive captions lacks corresponding negative captions (\(N\)=0), which does not meet the training requirements set for this task. To ensure the balance of the training data, we selected 1,000 complete samples from each sub-datasets which include both positive and negative captions to form a new training set. Accordingly, 100 samples from each subsets in validation set to create a new validation set, while testing set is not changed for a fair comparison.

In this paper, for training the Detic \cite{RN157} and OWL-ViT \cite{RN263} models, we utilize the corresponding loss functions from their original papers, which are the standard losses in a two-stage detector and a bipartite matching loss similar to that introduced by DETR respectively. However, since Grounding DINO \cite{RN141} is primarily a REC model, it differs from other detectors in that it cannot accept a vocabulary of captions as input. Therefore, following the inference approach in \cite{RN149}, we perform individual forward passes for each caption in the vocabulary and then calculate a bipartite matching loss for the predictions corresponding to the positive captions, similar to the loss used in OWL-ViT \cite{RN263}. Only the classification loss is computed for the negative captions without ground truth bounding boxes. For classification loss for both positive and negative captions, we adhere to the original Grounding DINO \cite{RN141} training method, setting the target binary classification label to 1 for each positive caption token and to 0 for the negative captions.
 
 \subsection{State-of-the-Art Comparison on FG-OVD Dataset}
 
Table \ref{tab:table2} shows the accuracy on the benchmark dataset before and after applying the HA-FGOVD method proposed in this paper, and comparing with previous state-of-the-art methods. Our proposed HA-FGOVD approach achieves an overall increase in accuracy when applied to each model. The Trivial subset exhibits minor fluctuations due to the fact that the positive and negative captions are completely dissimilar texts with inherently significant differences in their textual features. Consequently, the effect of enhancing attribute features to differentiate these textual differences is minimal. In individual sub-datasets, the proposed approach can achieve the highest accuracy improvement of up to 6.2\%. 
\footnotetext[1]{The authors corrected the value to 24.6 instead of 28, as indicated on the GitHub homepage of the paper \cite{RN149}. For further details, please refer to the official code repository: \url{https://github.com/lorebianchi98/FG-OVD}.}

Significantly, these experimental results show that the original text features of the OVD model indeed contain attribute features which are suppressed. By highlighting and enhancing attribute features and linearly combining them with global text features, attribute information can be activated, thus improving the detection ability of FG-OVD. After applying this method, the performance on multiple sub-datasets of the OWL-ViT model, as well as the average value of all subsets, reaches the state-of-the-art (SOTA) performance on the FG-OVD dataset.

\begin{table*}[!t]
	\setlength{\abovecaptionskip}{0.05cm}
	\centering
	\caption{ Overall ablation study on OWL-ViT(L/14) baseline for key factors of our proposed HA-FGOVD, including without bias, mask the first and last tokens ([CLS] and [SEP]) and and mask random words based on the proposed method, respectively. \textbf{Bolden} are the best results.} 
	\label{tab:tableablation}
	\begin{tabular}{cllllllll|l}
		\toprule
		Methods & Hard & Medium & Easy & Trivial & Color & Material & Pattern & Transp. & Average\\
		
			\midrule
			baseline   & 26.6 & 39.8 & 44.5 & 67.0 & 44.0 & 45.0 & 36.2 & 29.2 & 41.5\\
			w/o bias  & \textbf{31.6} \textcolor{red}{(+5.0)} & \textbf{46.0} \textcolor{red}{(+6.2)} &\textbf{50.9} \textcolor{red}{(+6.4)} & 67.0 (+0.0) & \textbf{48.6} \textcolor{red}{(+4.6)} & 48.3 \textcolor{red}{(+3.3)} & 36.9 \textcolor{red}{(+0.7)} & 30.3 \textcolor{red}{(+1.1)} & 45.0 \textcolor{red}{(+3.5)} \\
			mask \(\left[\mathrm{CLS}\right]\)\ \textbf{\text{\&}}\ \(\left[\mathrm{SEP}\right]\) & 29.0 \textcolor{red}{(+2.4)} & 43.0 \textcolor{red}{(+3.2)} & 45.3 \textcolor{red}{(+0.8)} & 67.1 \textcolor{red}{(+0.1)} & 46.6 \textcolor{red}{(+2.6)} & 47.3 \textcolor{red}{(+2.3)} & 35.1 \textcolor{green}{(-1.1)} & 31.6 \textcolor{red}{(+2.4)} & 43.1 \textcolor{red}{(+1.6)} \\
			mask random words  & 23.7 \textcolor{green}{(-2.9)} & 38.6 \textcolor{green}{(-1.2)} & 42.5 \textcolor{green}{(-2.0)} & 66.8 \textcolor{green}{(-0.2)} & 40.9 \textcolor{green}{(-3.1)} & 42.9 \textcolor{green}{(-2.1)} & 31.1 \textcolor{green}{(-5.1)} & 25.7 \textcolor{green}{(-3.5)} & 39.0 \textcolor{green}{(-2.5)}  \\
			HA-FGOVD (ours)  & 31.4 \textcolor{red}{(+4.8)} & \textbf{46.0} \textcolor{red}{(+6.2)} & 50.7 \textcolor{red}{(+6.2)} & \textbf{67.2} \textcolor{red}{(+0.2)} & 48.4 \textcolor{red}{(+4.4)} & \textbf{48.5} \textcolor{red}{(+3.5)} & \textbf{38.0}  \textcolor{red}{(+1.8)} & \textbf{32.7} \textcolor{red}{(+3.5)} & \textbf{45.4} \textcolor{red}{(+3.9)}\\
			
			\bottomrule
			
		\end{tabular}
	\end{table*}

\subsection{Transfer Evaluation of Explicit Weight Scalar Triplets}
To assess the transferability of the explicit weight scalar triplets among various OVD models, we transfer the optimal triplets from one model to another, and then evaluate their performance, as demonstrated in the Table \ref{tab:table3}. 

In the theory of deep learning, the bias parameter typically aiding in better data fitting of the models. Compared to the feature representation layers that capture universal features, such as the weights of convolutional layers, the value of bias is usually more specific to the particular model and task \cite{RN294}. Therefore, when testing the transfer performance of weight scalar triples, we set the bias to 0 for all. 

\begin{figure*}[!t]
  \centering
  \includegraphics[width=\textwidth]{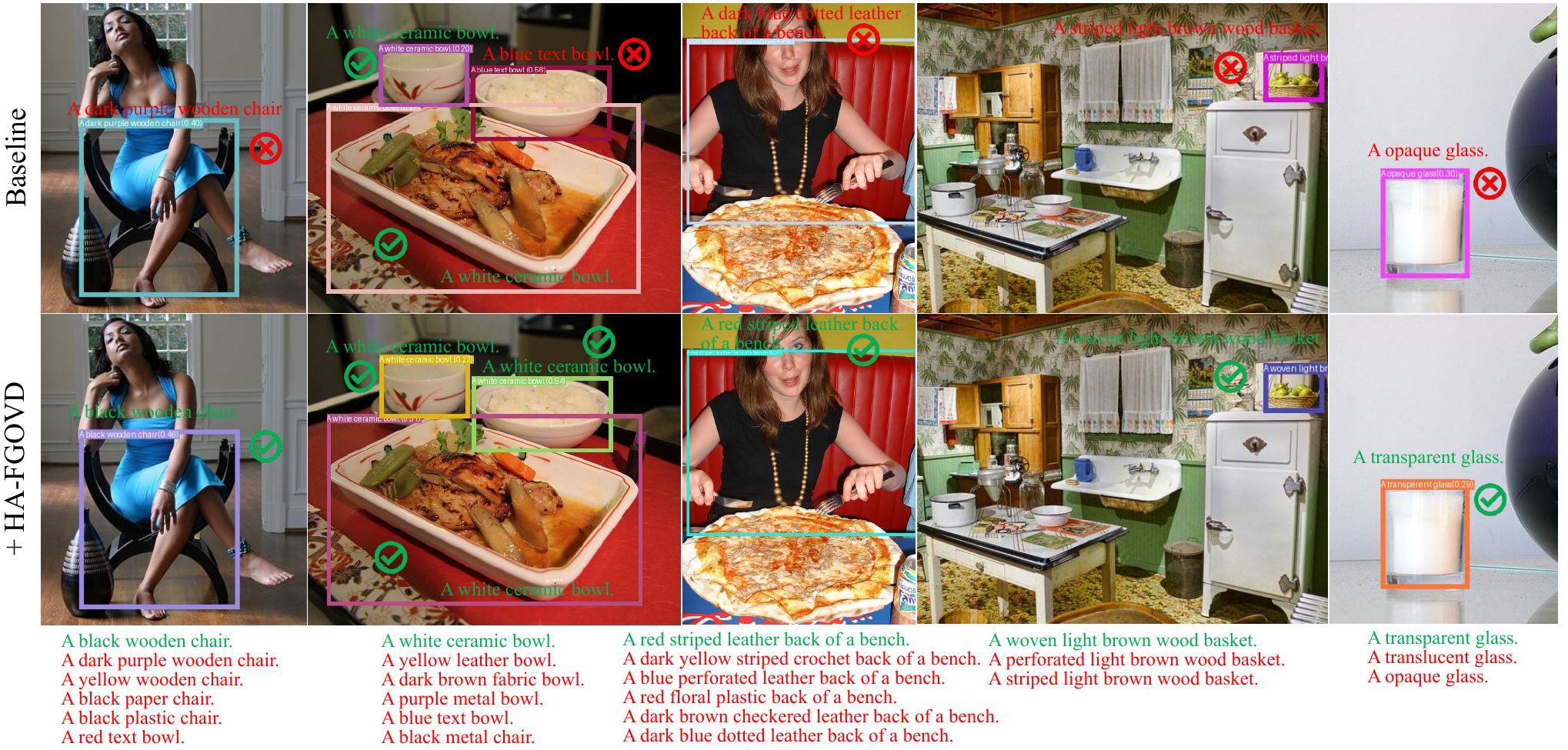}%
  \caption{The visualization of the detection results before and after applying our proposed HA-FGOVD approach on OWL-ViT baseline. At the bottom are the captions utilized for detection during the inference phase, with positive captions highlighted in green and negative captions in red. The initial row corresponds to the baseline results, which exhibit inaccuracies. The subsequent row delineates the refined results after applying our proposed HA-FGOVD approach, effectively corrects the aforementioned errors.}
  \label{fig_6}
\end{figure*}
Under the transferred parameter settings, the overall accuracy of the models can still increase. But the overall accuracy of the triplets obtained through training is more superior. Furthermore, the accuracy of several datasets surpasses that of the trained parameters. This indicates that when the training set is balanced, the parameters obtained by training are globally optimal. Otherwise, the models may reach a local optimum.

\subsection{Ablation Study}
We investigate the key factors for our method to work in Table \ref{tab:tableablation}. The ablation study is conducted with the best-performed OWL-ViT model under the same experiment settings.

The 1\textsuperscript{st} row in Table \ref{tab:tableablation} presents the mAP of the OWL-ViT baseline, while the last row illustrates the accuracy of our method. The 2\textsuperscript{nd} row delineates the results without the bias parameter in Eq. \ref{eq:bias}. Although non-bias setting still yields performance gain from the baseline, incorporation of the bias parameter leads to higher overall accuracy. This suggests that the bias can better fit all types of data by linearly translating attribute features, thereby achieving a global optimum.

To explore the significance of the initial and separator special tokens ([CLS], [SEP]) in extracted attribute features, the results of masking these tokens are reported in the 3\textsuperscript{rd} row. The data indicate that most sub-datasets exhibit an increase in performance; however, the accuracy of the pattern subset declines by 1.1\% compared to the baseline. In contrast, our proposed method, which does not mask the two tokens, demonstrates substantial improvements across all sub-datasets. This indicates that the special tokens in text encoders encompass vital fine-grained information that should not be masked.

Finally, to evaluate the LLM capability of highlighting attribute words, we conducted experiments by the random masking of certain input textual vocabularies, rather than attribute positions from LLM. Specifically, during both training and testing, we randomly selected \([0, L-1]\) vocabularies to mask within an input of \(L\) words, effectively augmenting \([1,L]\) words. In the 4\textsuperscript{th} row, mAP for all subsets decrease, indicating that attribute word extraction of LLM guarantees an improvement in fine-grained capabilities, thereby validating the effectiveness of our proposed approach.

\subsection{Visualization}

Fig. \ref{fig_6} provides the visualization results of OV detection for the baseline before and after applying the HA-FGOVD method. Our approach demonstrates excellent performance in improving the FG-OVD capability of mainstream OVD models, especially the complex input captions with a composition of various attribute types in 3\textsuperscript{rd} and 4\textsuperscript{th} columns in Fig. \ref{fig_6}, as well as captions that hard to distinguished from each other in 1\textsuperscript{st}, 2\textsuperscript{nd} and 5\textsuperscript{th} columns. In conclusion, our HA-FGOVD approach is robust and powerful enough to handle these challenging scenarios in FG-OVD task. 

\section{Conclusion}
In this paper, we propose a universal FG-OVD approach, called HA-FGOVD, which is a plug-and-play manner for mainstream frozen OVDs to enhance their fine-grained object detection capabilities by making use of a simple but explicit linear composition. Extensive experiments have demonstrated that our approach can effectively activate the suppressed attribute features, proving that there is potential for attribute features within the latent space of mainstream OVD models. Moreover, the weight scalar triplets in our approach can be transferred to other OVD models without additional training. The proposed approach achieves new state-of-the-art performance on the FG-OVD benchmark.

\bibliographystyle{IEEEtran}
\bibliography{IEEEabrv,reference}

%
%
%

%

\vfill

\end{document}